\documentclass[review,1p,authoryear]{article}

\usepackage{amsmath,amssymb,upgreek}

\usepackage[utf8x]{inputenc}

\usepackage[margin=1in]{geometry}

\usepackage[table]{xcolor}

\usepackage{array}

\usepackage{lastpage,fancyhdr,graphicx}
\usepackage{amsmath}
\usepackage{amsthm}
\usepackage{amsfonts}
\usepackage{amssymb}
\usepackage{graphicx}
\usepackage{mathalfa}
\usepackage{psfrag}
\usepackage[justification=centerfirst]{subfig}

\newcommand{\one}{($i$) }
\newcommand{\two}{($ii$) }
\newcommand{\three}{($iii$) }

\title{Constraints on Hebbian and STDP learned weights of a spiking neuron} % Please use "sentence case" for title and headings (capitalize only the first word in a title (o

\author{
Dominique Chu and Huy Le Nguyen\\
CEMS, School of Computing, University of Kent, CT2 7NF, Canterbury, UK
\\
dfc@kent.ac.uk
}

\begin{document}
\maketitle
\begin{abstract}
We analyse mathematically the constraints on weights resulting from  Hebbian  and STDP learning rules applied to a spiking neuron with weight normalisation.    In the case of pure Hebbian learning, we find that the normalised weights equal  the promotion probabilities of  weights  up to correction terms that depend on the learning rate and are usually small. A similar relation can be  derived for STDP algorithms, where the normalised weight values reflect a difference between the promotion and demotion probabilities of the weight. These relations are practically useful in that they allow  checking for convergence of Hebbian  and STDP algorithms. Another application is novelty detection. We demonstrate this  using the MNIST dataset.
\end{abstract}
%%%%%%%%%%%%%%%%%%%%%%%%%%%

%\begin{keyword}
%\end{keyword}

%\pagebreak
\section{Introduction}

Hebbian learning  \cite{hebbian} is a well established approach to train neural networks  that is not based on gradient descend. Instead, weights  are updated based on correlations between input and output.  There are a number of  features that make  Hebbian learning an attractive method to learn in the context of artificial intelligence  (AI).   For one, it is a   local rule and does not require the computation of gradients.  This makes it resource efficient and    suitable in situations where the computation of gradients is not practical  \cite{lammie}.  Secondly, it is biologically plausible and as such can be used to understand principles of learning in the brain. Thirdly,   and for  applications in AI perhaps  most importantly,   Hebbian learning  can be used as a part of supervised learning   \cite{stdpsym,boulet} but more importantly it  is a powerful method for  unsupervised learning  \cite{unsuper1,unsuper2}.
 For example, it   is  well known  that  under certain conditions the weight vectors of networks  trained with Hebbian learning rules  indicate the principal component vector of  the training  dataset  \cite{kingkong,oja}.  Given that the vast majority of all available data is unlabelled, this makes Hebbian learning an important part of the toolbox of AI.
\par
While Hebbian learning approaches  are not commonly used to train rate-coding neural networks, they have become an important approach   for training spiking neurons  \cite{hebbian}.  These are a type of  artificial neuron that    retains a ``memory'' of its state over time. This internal state is often called the ``membrane potential'' and typically is a real valued variable that decays over time.  Their  memory makes  spiking neurons naturally suitable for processing  temporal input data, such as video or audio. It is now well established that  a spiking neuron is a  more  powerful  information processors  than  the perceptron  \cite{maass,rubin}.  Another advantage of spiking architectures is that  they are biologically plausible  \cite{plausible} and  can be implemented   in energy efficient  special purpose hardware  \cite{neuromorphic,spinnaker,loihi}.  Implementations of spiking neurons including   the tempotron  \cite{tempotron} or the chronotron \cite{chronotron} and others  \cite{spikeapp1,spikeapp2} demonstrate in practice the abilities  of spiking neurons to learn  to classify spatio-temporal patterns  \cite{stdpsym,learnsignals} including  multi-label classification  \cite{gutig,mypaperwithfil}.  
\par
The focus of this article is to understand generic properties of weights that result from Hebbian training algorithms applied to a single  continuous time spiking neuron.  We shall probe various versions of Hebbian learning rules as well as the  asymmetric  {\em spike-timing dependent plasticity} (STDP) \cite{stdp} rule, where a Hebbian update is combined with an anti-Hebbian weight decrease for those weights that are active immediately after an output spike was generated. We also investigate how  different ways to normalise weights during Hebbian learning  \cite{normalisationsnn} impact on the allowed weights. 
\par
The most common way in the literature to model the  weights resulting from various training regimes is to identify the  eigenvalues of the weight update matrix   \cite{readinggroup1}. The problem with this approach is that it assumes that the weight evolution is a deterministic process. This is clearly not true because the  update dynamics of  weights  is normally  stochastic due to the random input.   An alternative is to use  master equation approaches   \cite{readinggroup1,stochneuralthesis,leanonmeold2} to capture stochastic aspects, but this is  rarely feasible. Approximations to the master equations, especially the Fokker-Planck equation are  common, but still  difficult to solve and in fact problematic  \cite{leanonme}. 
\par
Here, we take a different approach. Rather than attempting to derive the  distribution of weights, we  concentrate exclusively on the {\em metastable} and {\em absorbing states}, i.e. weight values that  the neuron will have {\em after} a long time of training and  {\em for} a long time. Our main result is a mathematical relation between the allowed metastable weight combinations that are consistent with Hebbian training and the probability of the weights to be increased/decreased during the Hebbian learning process. For a simple Hebbian learning rule we find that the probability of weight promotion  equals  the value of the normalised weight up to some correction terms that depend on the learning rate.
\par
The significance of this result is that it makes it possible to determine practically  whether or not a set of weights  is  adapted to the statistical bias of the input.  This is relevant for the following reasons:  Firstly, our results provide a novel characterisation of the weights that will evolve in the long run during Hebbian learning. This is interesting in itself, but will hopefully also open up novel pathways for the mathematical analysis of the behaviour of spiking neural networks trained using Hebbian learning. In particular, it  may lead to novel insights about the   statistical distribution of weights, which  has been of substantial research interest   \cite{weightsinnns,schwabweights,weightsinlif,inthebrain}. Secondly and more practically relevant, as we discuss below, our results can be used to adapt the learning rate of the neuron. In many contexts it is beneficial to have a large learning rate at the beginning and a small or vanishing learning rate when the neuron is close to what it should learn. The relationships we derive make it possible to estimate how far the neuron is away from ideal weights. The learning rate can be decreased accordingly. Finally, as we will show below, the relationships we derive can also be used for novelty detection.

\section{Results}

In this article, we will consider a number of variants of Hebbian learning. The basic idea of Hebbian learning is best summarised by the well known tenet ``What fires together, wires together.'' In the context of an artificial neuron whose input channels are weighted this would mean  that the activity of an input channel can lead to an increase of the weight of the input channel if it happens within a short temporal window before the neuron itself spikes. There is then a sense of the input ``causing'' the output  spike of the neuron. 
\par
This basic Hebbian learning rule is asymmetric in the sense that only activities before the output spike matter, and the only option for weight changes is to increase the weights (although this is typically paired with some mechanism to prevent unlimited weight growth). This basic Hebbian scheme  is frequently  supplemented  by a rule that decreases the weights of input channels firing {\em after} an output spike happened. This {\em asymmetric}  rule is usually referred to as  {\em spike timing dependent plasticity} (STDP) rule.    
\par
For both the symmetric and asymmetric rule there are many variants in the literature.  For example, the amount of weight increase can depend on the precise distance from the output spike, i.e. the closer the input spike to the output spike the greater the decrease/increase of the corresponding weights. Furthermore, this dependence could be linear or non-linear. 
\par
Below, we will first concentrate on  asymmetric (Hebbian) rules. We will model a number of variants of these rules, including different normalisation schemes. As a main result, we will find that in steady state  the weights of a channel are directly related to the probability of the channel being promoted. We then extend this to the case of symmetric rules, where we find a similar albeit slightly more complex relationship between weights and promotion probabilities.

\subsection{Model of the spiking neuron}

Throughout this contribution we model the  spiking neuron as having a time-dependent internal state  $V\in\mathbb{R}^+$ --- the ``membrane potential.'' The neuron receives discrete input spikes in continuous time  each arriving  via one of  $N$ different channels; these are modelled as  instantaneous  spikes $\delta(t-T_k^i)$  arriving  at channel $i$ and  time $T_k$ where $t,T_k\in \mathbb{R}^+$;  $\delta(\cdot)$ denotes the Dirac delta function. Each input channel $i$ is associated with a weight $w_i$, such that at spike time $T_k^i$ the membrane potential $V$ increases instantaneously  by $w_i$. The membrane potential also decays (or ``leaks'') in time with a rate constant $d\geq 0$, such that the instantaneous rate of decay at time $t$ is $V(t) d$.   The neuron also has a spiking threshold. If $V$ crosses the threshold from below, then  \one  the membrane potential is reset to 0 \two the weights are adjusted according to one of the learning rules described below. Crossing of the threshold is normally  also associated with the generation of an output spike. Since we only model a single neuron, output spikes will be of no further relevance here.

\subsection{Relationship between promotion probability and weights}

\subsubsection{The long-term behaviour of the weight vector}

The  simplest  way to  derive the steady-state values of   weights  $w_1, w_2,\ldots,w_N$ of a neuron is to analyse the fixed points of the weight update rule $f(x_1,\ldots,x_N,w_1,\ldots,w_N,y,t)\overset{!}{=}0$, which is a function of the inputs $x_1,\ldots,x_N$ to the neuron as well as its output $y$.
\begin{equation}
\Delta  w_i =   f(x_1,\ldots,x_N,w_1,\ldots,w_N,y,t)
\end{equation}
This function $f$ could, for example, be the famous {\em Oja rule}, i.e. 
\begin{displaymath}
f(x_i, w_i,y)=  x_i\cdot y - y^2\cdot w_i
\end{displaymath}
In steady state, updates would leave the weight vector unchanged, i.e.  on average
\begin{equation}
x_i\cdot y = y^2\cdot w_i
\end{equation}
This update rule can be shown to lead to a fixed point where the weights of the neuron align with the principal component vector of the input data if the input data  has a vanishing mean \cite{readinggroup1}.  In general, Hebbian updates of the weight vector  rules will not compute the  principal component  of the input function, but compute some other functions. In any case, for the steady state weight vector it will always be the case that $f(x_i, w_i,y)= 0$ for all $i$.
\par
A slightly different perspective on this is possible. A Hebbian update rule that avoids unlimited growth of weight values requires an update rule that contains both weight increases according to  some rule $f^+= f^+(x_1,\ldots,x_N,w_1,\ldots,w_N,y,\epsilon)$  but also weight decreases $f^-= f^-(x_1,\ldots,x_N,w_1,\ldots,w_N,y,\epsilon)$, where  $f^+,-f^-\geq 0$. Any particular  update event  is either positive ($f^+$) and leads to a weight promotion  or negative ($f^-$) effecting a weight demotion. A quantity of interest is the average  of the  step size $s_i:= \Delta w_i$ during an update event given by 
\begin{equation}
\label{masterpup}
\langle s_i\rangle := p_i f^+ + (1-p_i ) f^-,
\end{equation}
 where $p_i$ is the (unknown)  probability that the $i$-th weight is promoted. A necessary  condition for weights to stabilise is $\langle s_i\rangle=0$. Formally, we can now solve the  right hand side of Eq. \ref{masterpup}  for $p_i$ to obtain a relationship between the promotion probability and the value of the weights.
\par
We now consider the case of a  simple Hebbian learning rule for a spiking neuron in continuous time with weight normalisation. The scenario we consider is as follows:
\begin{enumerate}
 \item 
Whenever the neuron receives an input from channel $i$ then the membrane potential is increased by $w_i$.
\item
If the membrane potential exceeds a critical threshold value $\theta$, then an output spike is generated and the membrane potential is reset to 0.
\begin{enumerate}
 \item 
If the  $k$-th input channel  was active at least $\uptau$ time units before an output spike was generated, then the $k$-th weight is increased by some fixed amount $\epsilon$ --- the {\em learning rate}.
\item
Following an update step weights are normalised  by setting $w_i \rightarrow w_i/\left(\sum_j w_j^l  \right)^{1\over l}$, for some real number $l\geq 1$ and all $i=1,\ldots,N$.   
\end{enumerate} 
\end{enumerate} 
\par
It is now straightforward to derive a  relationship between $w_i$  and the probability $p_i$  that the $i$-th weight will be increased during a particular promotion event.   The simplest case is  $l=1$, i.e. when the weights are always normalised to 1 following an update event. This case also coincides with Oja's rule. The average step size can then be written as follows:
\begin{equation}
\label{balance}
\langle s_i\rangle = p_i \left({w_i + \epsilon\over 1+\epsilon} - w_i \right) - (1-p_i) \left( w_i -{w_i \over 1+\epsilon}\right).
\end{equation}
 The first term on the rhs of this equation simply accounts for the  increase of weight $i$ by $\epsilon$ followed by a weight normalisation which leads to a slight decrease again. The net effect is always an increase, which happens with probability $p_i$. 	 The second term describes that  during  any given weight update event, it may also be the case that any of the other $j\neq i$ weights are increased instead. In this  case the weight is normalised without being increased. This leads to a net decrease of the $i$-th weight. Note that the probabilities to be increased and decreased are typically a  function of the weights themselves.  Solving Eq. \ref{balance} for $p_i$, we obtain the condition:
\begin{equation}
\label{simplestcase}
p_i= w_i 
\end{equation}
%
%`
This means that, after a sufficiently long training period the weights will be equal to the promotion probability. In the limiting case of $\uptau\to 0$  the weight would be the probability that the corresponding channel fired last before an output spike was triggered by the neuron.
\par
The same procedure, although with a bit more notational effort, can be applied to the case of $l>1$. In this case the normalisation factor depends on which weight was promoted  and     Eq. \ref{masterpup} becomes:
\begin{equation}
\label{gennorm}
\langle s_i\rangle = p_i \left({w_i + \epsilon \over \varsigma_i } - w_i \right) 
+ 
\sum_{j\neq i} p_{j}\left(  {w_i  \over \varsigma_j} - w_i  \right),
\end{equation}
where $\varsigma_i:= (w_1^l + \cdots  +w_{i-1}^l + (w_i+\epsilon)^l + w_{i+1}^l + \ldots+ w_N^l)^{1\over l} $ and  $p_{j}$ is the probability that the weight  of the $j$-th channel is promoted.
\par
From Eq. \ref{gennorm}  we can now establish a relation between the promotion probability and the weights
\begin{align}
& p_i w_i\left({1\over{\varsigma_i}} -1\right) + p_i {\epsilon\over \varsigma_i} +
 w_i\sum_{j\neq i} p_j   \left({1\over{\varsigma_j}} -1\right) =0\nonumber \\
& p_i  {\epsilon\over \varsigma_i} = w_i\left(1-\sum_{j} p_j   {1\over \varsigma_j}\right)\nonumber\\
& {p_i\over \varsigma_i} = {w_i\over \epsilon}\left(1 - \left\langle {1\over \varsigma}\right\rangle\right). 
\label{firstrel}
\end{align}
The rhs of this equation is still somewhat opaque in that the average over the inverse normalisation factors $\varsigma_i$ is dependent on $w_i$ and $p_i$ such that further analysis is necessary.  In order to evaluate this average it  is useful to write  $\varsigma_i= (\mathcal W + (w_i+\epsilon)^l -w_i^l))^{1\over l}$, where   $\mathcal W := \sum_i w_i^l$. For a $l$-normalised weight vector we have $\mathcal W=1$.   Since $\epsilon$ is a small term, we can expand the expression to first order in $\epsilon$, which yields:
\begin{align}
\varsigma &=
\exp\left(   {\frac {\ln  \left( \mathcal W+{ \exp\left(l\ln  \left( w_i \right) \right) }-{
w_i}^{l} \right) }{l}} \right)-
{\frac 
{   \exp\left({\frac {  l \ln(w_i) +   \ln  \left(\mathcal W +{\exp\left(l\ln  \left( w_i \right) \right)}-{w_i}
^{l} \right) }{l}}\right)  }
{w_i
 \left( \mathcal W +{\exp\left(l\ln  \left( w_i \right) \right)}-{w_i}^{l} \right) }
}\epsilon
+
O \left( {\epsilon}^{2} \right) 
\nonumber \\
&\approx {\mathcal W^{1+l\over l}w    + \exp\left({l^2\ln(w_i) + \ln(\mathcal W)\over l}\right)\epsilon
\over
\mathcal W w_i
 }\nonumber\\
&=\mathcal W^{1\over l}+ w_i^{l-1} \mathcal W^{1-l\over l}\epsilon =  1+ w_i^{l-1} \epsilon
\label{normterm}
\end{align}
Next, we evaluate the average over the inverse  normalisation term $\left\langle{1\over \varsigma}  \right\rangle$. First,  we take note of the general relation
\begin{displaymath}
{1\over A + x} = {1\over A} - {1\over A^2}x + O(x^2),
\end{displaymath}
and expand the average to first order in the second term:
\begin{align}
\left\langle{1\over \varsigma}  \right\rangle &= \sum_i p_i {1\over 1 + w_i^{l-1} \epsilon}
\nonumber\\
&\approx \sum_ip_i\left( 1  -   w_i^{l-1}\epsilon\right)
\nonumber \\
&=     1 -   \sum_i p_i w_i^{l-1}  \epsilon
\nonumber\\
&=   1- {\langle w^{l-1}\rangle} \epsilon 
\label{avgnormterm}
\end{align}
Substituting this back into Eq. \ref{firstrel} and re-arranging the terms we find the first order expression of the promotion probability in terms of weight:
\begin{equation}
p_i= { \langle w^{l-1}\rangle \over 1-w_i^{l-1}\epsilon} w_i
\label{asymcase}
\end{equation}
This expression is compact, but contains implicitly $p_i$ on the right hand side. Taking advantage of the fact that the promotion probability is normalised, we can establish that the rhs of the equation is the normalised weight, thus obtaining
\begin{equation}
p_i=  {w_i \over  \sum_j w_j} \qquad\qquad   \sum_jw_j = { \left(  1-\langle w_i^{l-1}\rangle\epsilon \right) \over \langle w^{l-1}\rangle}.
\label{symmcase}
\end{equation}
Note that in deriving this latter equation, we did not make any explicit assumptions about the length $\uptau$ of the time window. This information is implicit in the $p_i$ in that the meaning of $p_i$ changes as $\uptau$ is adjusted. While the length of the time window will clearly affect the weights that the system will find, it does not change the constraint that the normalised weights equal the promotion probability.

%----

\subsection{Symmetric STDP}

In practical applications  of spiking networks, the basic Hebbian learning rule is often augmented by  an additional anti-Hebbian term.   In STDP rules   weights are increased when the corresponding input fires before an output spike is generated and they are decreased when the input fires shortly after an output spike.  We assume, for the moment,  that the change of the weights is  $\pm\epsilon$ if the input channel fires within a period of  $\pm\uptau$ time units of  an output spike. 

\par
This case can be modelled largely along the same lines as above, but some additional notational complexities  become necessary.  The  expression for the average step size $\langle s_i\rangle$ of the weight $w_i$ needs to be extended to take into account the anti-Hebbian component. For simplicity, we assume that the Hebbian increases/decreases are by  $\pm\epsilon$, although the results to follow can be adjusted easily for non-equal changes. 
\par
 In analogy to equation \ref{gennorm}, we can write the average step size:
\begin{align}
\langle s_i\rangle &= k^\mathrm{p}\left[ w_i p_i \left({1\over{\varsigma_i^\mathrm{p}}} -1\right) +p_i{\epsilon\over \varsigma_i^\mathrm{p}} +w_i \sum_{j\neq i} p_j \left({1\over{\varsigma_j^\mathrm{p}}} -1\right) \right]
\nonumber \\
\label{intem2}
&+k^\mathrm{d} \left[ w_i  q_i \left({1\over{\varsigma_i^\mathrm{d}}} -1\right) - q_i{\epsilon\over \varsigma_i^\mathrm{d}} +w_i \sum_{j\neq i} q_j \left({1\over{\varsigma_j^\mathrm{d}}} -1\right) \right]
\end{align}
Here the first term in the square brackets is identical to Eq. \ref{gennorm} and formulates the contribution to step sizes as they arise from the weight promotion events. The  unconditional probability of a weight promotion is  $k^\mathrm{p}$ and the conditional probability $p_i$ that there is a promotion event and that  the $i$-th channel is promoted. The second term takes into account weight demotions which happen  with  the unconditional   probability $k^\mathrm{d}$;   $q_j$ denotes the conditional probability that given a weight is demoted it is that of the   $j$-th input channel.
\par
A few clarifying remarks are appropriate before continuing. Firstly, the definition of   $k^\mathrm{p}, k^\mathrm{d}$ implies the normalisation condition $k^\mathrm{p} + k^\mathrm{d}=1$.  Secondly, $q_i$ and $p_i$ are conditional probabilities, but for  notational simplicity, we will henceforth refer to them as probabilities throughout the rest of this article.  Finally, note that   in Eq. \ref{intem2} it is necessary to distinguish between 2 different normalisation terms, that is  $\varsigma_i^\mathrm{p}$, which is the same as above and   $\varsigma_i^\mathrm{d} :=(w_1^l +  w_2^l + \ldots + w_{i-1}^l + (w_i-\epsilon)^l + w_{i+1}^l + \ldots )^{1\over l}$, which  arises from the demotion of weights. Note also that in Eq. \ref{intem2}  the minus sign in front of the second term in the second line comes from the fact that in the case of a demotion the weight change equals $(w_i - \epsilon)/\varsigma_j^\mathrm{d} - w_i$.
\par
To proceed  we  manipulate  Eq. \ref{intem2} in analogy to the Hebbian case  to find an expression for the average step size.
\begin{equation}
\langle s_i\rangle =k^\mathrm{p}w_i\left(\left\langle{1\over \varsigma}  \right\rangle_\mathrm{p} -1\right)
+
k^\mathrm{d}w_i\left(\left\langle{1\over \varsigma}  \right\rangle_\mathrm{d} -1\right) 
+
k^\mathrm{p} p_i  {\epsilon\over \varsigma_i^\mathrm{p}}
-
k^\mathrm{d} q_i    {\epsilon\over \varsigma_j^\mathrm{d}}
\end{equation}
Here, $\langle\cdot\rangle_\mathrm{d}$ denotes the average over all demotion events and $q_i$ is the probability that channel $i$ is demoted. In steady state, we then obtain the following relation for the promotion probability:
\begin{equation}
\label{pcalc}
{p_i\over \varsigma_i^\mathrm{p}} -
{k^\mathrm{d}\over  k^\mathrm{p}}  {q_i\over \varsigma_i^\mathrm{d}  } =
 -w_i \left( { k^\mathrm{p}\left(\left\langle{1\over \varsigma}  \right\rangle_\mathrm{p} -1\right)
-
k^\mathrm{d}\left(\left\langle{1\over \varsigma}  \right\rangle_\mathrm{d} -1\right)
\over
k^\mathrm{p}\epsilon
} \right)  
\end{equation}
We can now substitute in the first order approximations for the normalisation terms Eqs. \ref{normterm} and \ref{avgnormterm}. Note that these are different for the down and up probabilities in that the  sign of the first order correction is swapped. This gives us:
\begin{equation}
{k^\mathrm{p}\over  \varsigma_i^\mathrm{p}} p_i     -
{k^\mathrm{d}\over  \varsigma_i^\mathrm{d}}  q_i   
=
 w_i 
\left( { k^\mathrm{p}\left\langle w^{l-1} \right\rangle_\mathrm{p} 
-
k^\mathrm{d}\left\langle  w^{l-1}   \right\rangle_\mathrm{d} 
} \right)  
\label{notyetfinished}
\end{equation}
Here,  the learning rate has apparently fallen out of the equation. It  will be re-introduced  by substituting  in  the approximation of   the normalisation factors, which now become first order correction terms. Using once again Eq. \ref{normterm}, we obtain
\begin{equation}
k^\mathrm{p}p_i  (1 -  w^{l-1}_i\epsilon)
-
k^\mathrm{d}q_i  (1 +  w^{l-1}_i\epsilon)
=
 w_i 
\left( { k^\mathrm{p}\left\langle w^{l-1} \right\rangle_\mathrm{p} 
-
k^\mathrm{d}\left\langle  w^{l-1}   \right\rangle_\mathrm{d} 
%\over
%k^\mathrm{p}
} \right).  
\end{equation}
As before, we calculate  the averages in the second term on the rhs by exploiting that both probabilities must sum to 1.
\begin{equation}
\sum_j w_j=
{
k^\mathrm{p}  (1 -  \langle w^{l-1}\rangle_\mathrm{p}\epsilon)
-
k^\mathrm{d}  (1 +  \langle w^{l-1}\rangle_\mathrm{d}\epsilon)
\over
k^\mathrm{p}\left\langle w^{l-1} \right\rangle_\mathrm{p} 
-
k^\mathrm{d}\left\langle  w^{l-1}   \right\rangle_\mathrm{d} 
%\over
%k^\mathrm{p}
}
. 
\label{normcond} 
\end{equation}
With this, we obtain a relationship between the promotion/demotion probabilities and the weights.
\begin{equation}
{k^\mathrm{p}p_i(1 - w^{l-1}_i\epsilon) 
-
k^\mathrm{d}q_i(1 + w^{l-1}_i\epsilon) 
\over
k^\mathrm{p}(1-\left\langle w^{l-1} \right\rangle_\mathrm{p} )
-
k^\mathrm{d}(1+\left\langle  w^{l-1}   \right\rangle_\mathrm{d} )
}
   =
{ w_i\over \sum_j w_j}    
\label{epsilonconstant}
\end{equation}
This tells us that up to corrections in the learning rate the normalised weights equal a difference between the {\em unconditional}  promotion probability  $k^\mathrm{p}$  and demotion probability  $k^\mathrm{d}$.  In the case of $k^\mathrm{p}=1$ and $k^\mathrm{d}=0$ this reduces to the relation we found for the pure Hebbian learning rule  Eq. \ref{asymcase}, as it should.   
\begin{figure}
\centering
 \includegraphics[angle=-90,width=0.45\textwidth]{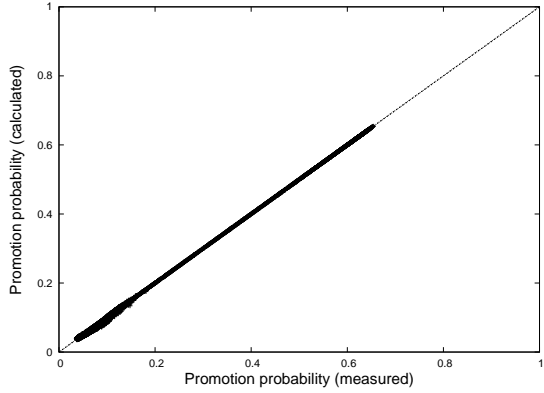}
 \caption{Comparing  $p_i$, calculated using Eq. \ref{pcalc}  with the estimated value from the simulation. This is an example with 10 input channels,  $\epsilon=0.0001$ and a $\uptau =0.07$. Each input channel fires randomly with a rate of $0.9$. The dashed line indicates the diagonal. }
 \label{stdpexample}
 \end{figure} 
\par
In practical applications of STDP the update step is often chosen to be non-constant. A common model choice is  to increase/decrease the weights that fire immediately before/after   the output spike most, and much less those input channels that were active after the output spike. The update rule is normally determined by  some function $\epsilon(\tau)$ for example  a truncated  exponential function  to determine the weight increases/decreases. 
\par
In order to model this most general case a straightforward, if notationally tedious extension of Eq. \ref{intem2} is necessary. The main change is that the step size now becomes a function of time since the last spike, $\epsilon=\epsilon(\tau)$ and $-\uptau\leq\tau\leq\uptau$. Eq. \ref{intem2} needs to be adjusted accordingly by integrating over  all possible spike times $\tau$, such that, for example, the first term becomes,
\begin{equation}
 k^\mathrm{p}\left[ w_i \sum_{j\in\pi_i} \int_{-\uptau}^0 p_j f_j(\tau) \left({1\over{\varsigma_j^\mathrm{p}(\tau)}} -1\right)d\tau\right].
\end{equation}
Here $f_j^\mathrm{p}$ is a probability density defining the spike time distributions of input channel $j$ before an output spike is generated.  The normalisation condition on the probabilities holds, i.e. 
\begin{displaymath}
 \sum_j   \int_{-\uptau}^0 p_j f_j^\mathrm{p}(\tau) d\tau  =1 .
\end{displaymath}
If we now  define an average learning rate  specific to the index $i$ as  $\langle\epsilon\rangle_{\mathrm{d},i}:=  \int_{-\uptau}^0  f_i^\mathrm{p}(\tau)\epsilon(\tau) d\tau$
then we can perform the analogous analysis  to arrive at the same expression as Eq. \ref{pcalc} to relate the weights and the promotion probability. 
\begin{equation}
p_i= q_i{k_\mathrm{d} \over k_\mathrm{d}}
{
\langle\epsilon\rangle_{\mathrm{d},i} - \langle\epsilon^2\rangle_{\mathrm{d},i} w^{l-1}
\over
\langle\epsilon\rangle_{\mathrm{p},i} - \langle\epsilon^2\rangle_{\mathrm{p},i} w^{l-1}
}
+
w_i
{
k_\mathrm{p}   \langle w^{l-1}\epsilon\rangle_\mathrm{p}          -  k_\mathrm{d} \langle w^{l-1}\epsilon\rangle_\mathrm{d}
\over
k_\mathrm{p}
(\langle\epsilon\rangle_{\mathrm{p},i} - \langle\epsilon^2\rangle_{\mathrm{p},i} w^{l-1})
} 
\end{equation}
Here the order of terms is somewhat obscured by the fact that average learning rates appear in all terms. Note, that the  terms that are linear in $\langle\epsilon\rangle$ are actually of order zero, whereas the quadratic terms are of first order. Keeping this in mind  it is straightforward to see that the case of $\epsilon(\tau)=\mathrm{constant}$ reduces to equation \ref{epsilonconstant}. Finally, in the case of pure Hebbian learning $k_\mathrm{d} =0$ and $k_\mathrm{p} =0$, we find again that the normalised weights equal the promotion probability, yielding in zero-th order:
\begin{equation}
p_i= 
{
 \langle w^{l-1}\epsilon\rangle_\mathrm{p}          
\over
\langle\epsilon\rangle_{\mathrm{p},i} 
} {w_i}. 
\end{equation}

\subsubsection{Decay model}

Weight normalisation is not the only way to maintain finite weights in  Hebbian learning rules. Another method is weight decay. The idea  is that in addition to weight increases due to the Hebbian learning rule, weights decrease (or ``decay'')  continuously according to some rule.  There are numerous possible  ways in which  weight decay could be implemented. Most of them are much simpler to analyse than the weight normalisation. Here, we only hint at possible results by way of a specific model. Concretely  we assume a setting with asymmetric STDP with some window length $\tau$  and a constant $\epsilon$. During each promotion event all weights are  decreased by a constant fraction $\delta$ of the weight value  and then increased by a fixed amount of $\epsilon$. The average step size is then much easier to write down than in the normalisation case. 
\begin{equation}
\langle s_i\rangle =  p_i (\epsilon - \delta w_i)  - (1-p_i)\epsilon
\label{stdpdecay}
\end{equation}
It is trivial to find an exact steady state solution to this.
\begin{equation}
\label{decayeq}
 w_i   =   {\epsilon\over \delta}   p_i
\end{equation}
Hence, again, in the case of Hebbian learning combined with weight decay, the normalised weights equal the promotion probability and the weight normalisation factor is given by $\sum_j w_j={\epsilon\over\delta}$.

\subsection{Some interesting examples}

We now illustrate the above results using  numerical examples of Hebbian trained neurons. For most examples below, we shall use unbiased random input data, but we also confirm the constraints on biased input data. We also demonstrate novelty detection using the MNIST data. All experiments below were conducted on  an Intel i7-6700 CPU  with 32 GB RAM.

\subsubsection{2 weights only}

As the  first example, we probe the simplest possible case of a  neuron with two input channels, $l=1$ and trained with an asymmetric Hebbian learning rule. While of no clear practical use, this case is interesting because it is possible to calculate the promotion probability without simulating the neuron. To do this, we enumerated all possible input sequences of length 13 (which was sufficiently long to capture all cases). These cases are:
\begin{enumerate}
 \item 
All 13 firing events are from input channel 1.
\item
The first 12 firing events are from input channel 1, but the last one from input channel 2.
\item
The first 11 firing events are from input channel 1, the 12$^\textrm{th}$ from input channel 2, the 13$^\textrm{th}$ from input channel 1
\item
etc\ldots
 \end{enumerate} 
 Having generated all possible input sequences, we  then generated waiting times between all input spikes by sampling from an exponential distribution with appropriately chosen rate parameter.  For each set of sequences thus generated, we then determined the point of the sequences where the membrane potential first exceeds our (arbitrarily)  chosen  threshold of $0.94$ and truncated the sequence   at this point. This left us with  the set of all  possible paths to crossing the threshold. 
\par
\begin{figure}
\centering
\subfloat[][]{ \includegraphics[angle=-90,width=0.45\textwidth]{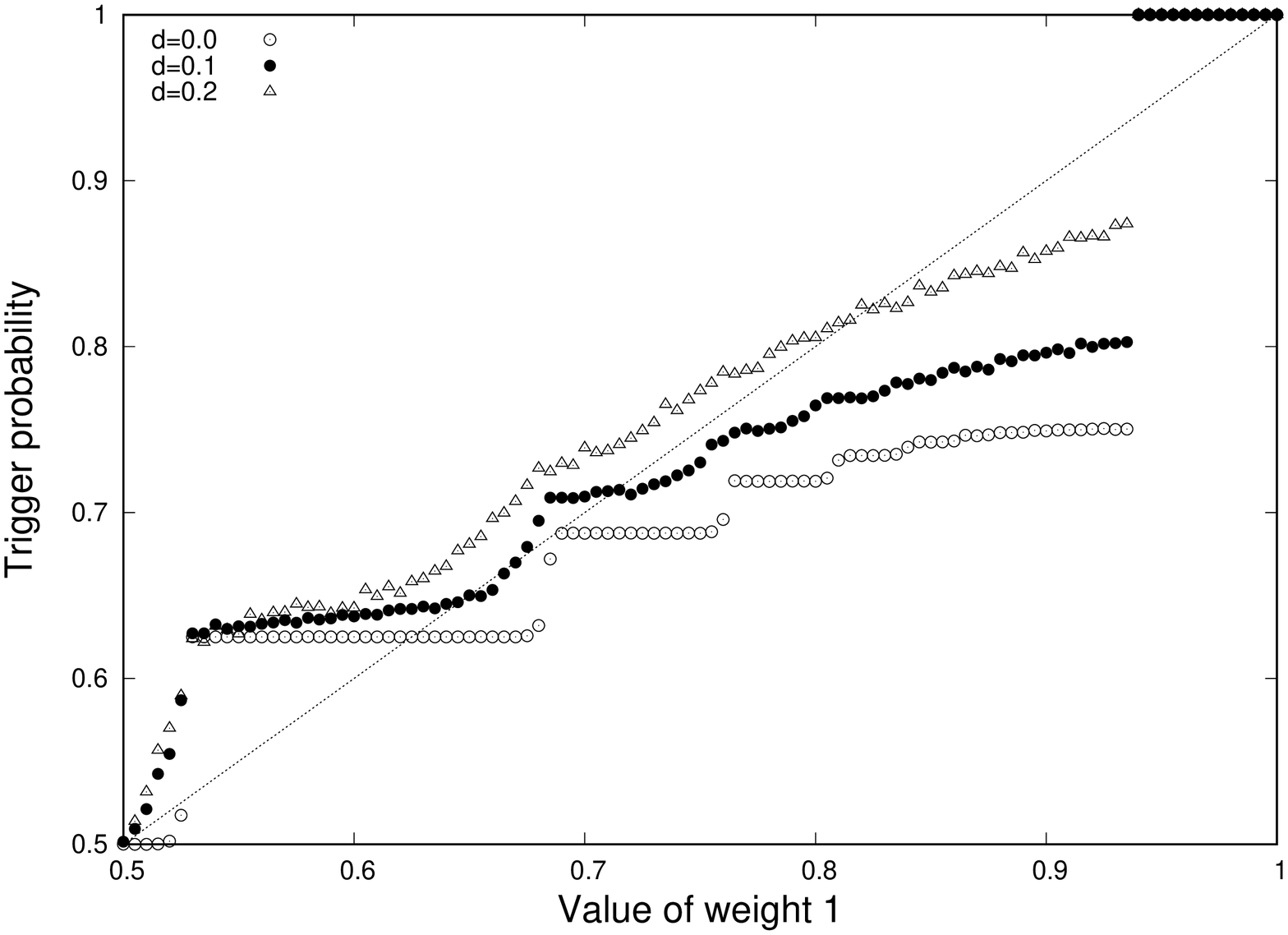} \label{fracs}}
\subfloat[][]{ \includegraphics[angle=-90,width=0.45\textwidth]{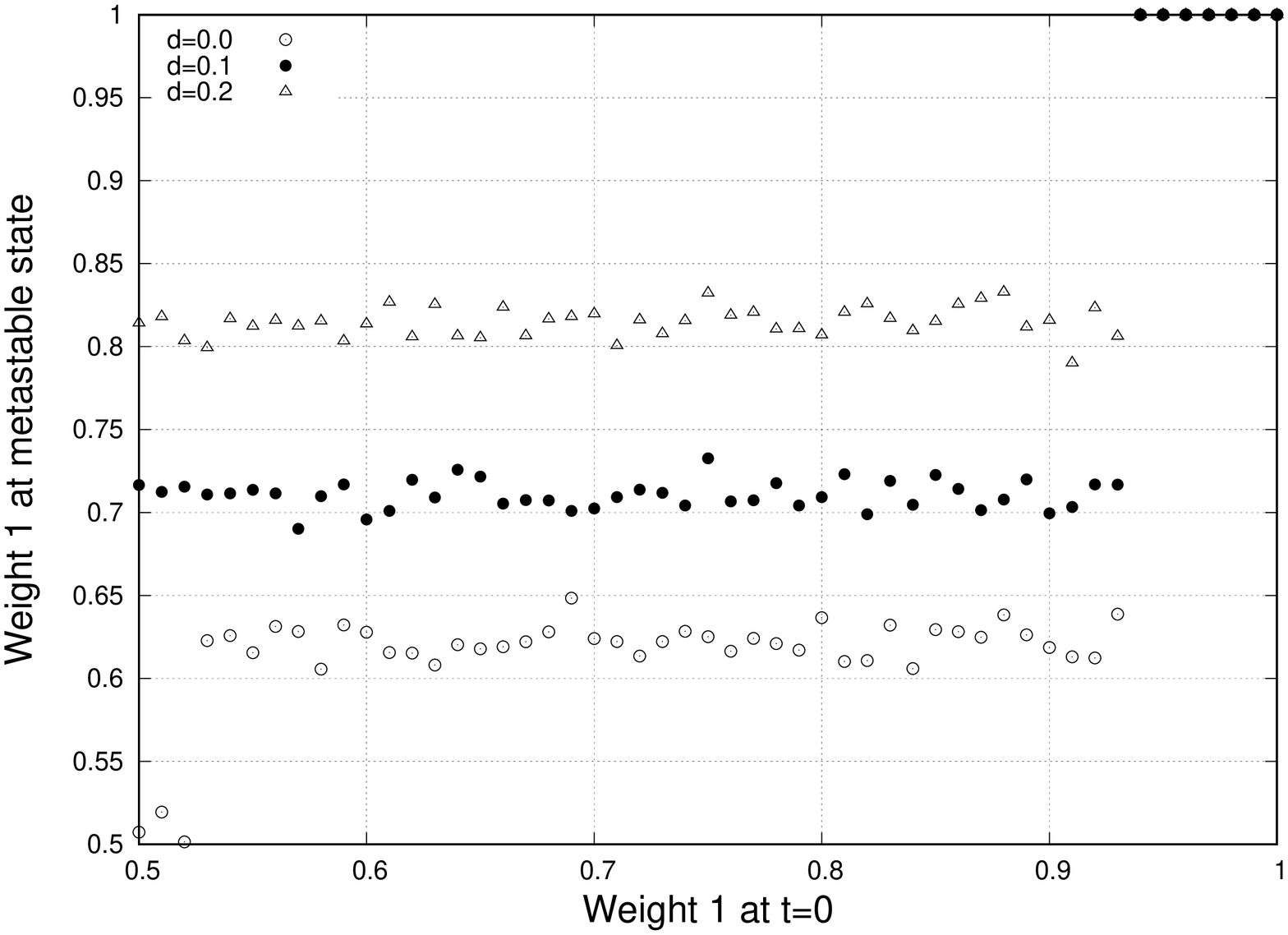}\label{w2dec}}
 \caption{\protect\subref{fracs} Assuming a neuron with 2 input weights and a threshold set to 0.94 and $l= 1$. The figure shows $p_1$ as a function of $w_1$ calculated explicitly as described in main text  for three different decay rates of membrane potential ($d=\{0.0,0.1,0.2\}$). \protect\subref{w2dec} Weights of the neuron after Hebbian training for 300000 time units  assuming a step size of 0.0005. Each point corresponds to a single simulation/calculation in both figures. The $p_i$ values are limited to $p_i$ values where the corresponding graph in \protect\subref{fracs} crosses the diagonal.}
 \label{example}
 \end{figure}

Given all truncated sequences, we then counted the number of instances where the sequence was truncated at weight 1, i.e. where the output spike was triggered by the first channel. Then we divided this number by the total number of sequences to obtain $p_1$.  This  probability  depends on the weights.  For example,  trivially $p_1=0$ if $w_1=0$ and similarly, it must be  $p_1=1$ for $w_1=1$. In-between these two extreme values $p_1$ can only be calculated by exhaustive enumeration of all possible sequences; see Fig. \ref{fracs} for a numerical example of $p_1$ vs $w_1$. Note that due to the method we use, the results we generated depend on the random seed,  in that repetitions will give slightly different inter-spike intervals each time. However, we found the variation between seeds to be negligible.
\par
A consequence of   Eq. \ref{simplestcase}  is that the weights of a  trained  neuron  are limited to the values  where $p_i$ intersects the diagonal in Fig. \ref{fracs}.   To check this, we initialised a neuron with many different values of $w_1$ and $w_2=1-w_1$ and trained it using Oja's rule using an  input  spiking rate  of $0.9$ for both channels  and  a  threshold of $\theta=0.94$. We then recorded the weight of input channel 1 after 300000 time units as a function of the initial weight; see Fig. \ref{w2dec}. We restricted the analysis to $w_1 \geq 0.5$. The rest  of the weights space can be inferred by symmetry.      
\par
In the case of no decay of the membrane potential, i.e.  $d=0$, Fig. \ref{fracs}  suggests  that there are only  4  weight values that meet the constraint $p_1=w_1$. Out of those 4 values, however,  we observed only 3 in the Hebbian trained neuron; see Fig. \ref{w2dec}: \one  For low initial values the weights remain symmetrically distributed with an average of $1/2$ for each of the channels. \two The second stable value is $\approx 0.63$. \three  Very high initial weights lead to a convergence to $w_1=1$. The fourth point  with a vanishing step size, at around $w_1 \approx 0.68$, is not taken by the system. 
\par
 The standard interpretation of this is  that $w_1 \approx 0.68$ is an ``unstable fixed point.''  Indeed,  a dynamic instability in Fig. \ref{fracs} exists if (when read from from left to right) $p_1$ crosses the diagonal from below. In this case, the weights will  be driven away from the fixed point.  On the other hand, if $p_1$ crosses the diagonal from above, then the weights will be driven towards the stable state. Following this reasoning, it is understandable why the system  omits one of the possible steady state weight values. However, there is more to this.
\par
If we increase the decay rate of the membrane potential, i.e. $d>0$, then  we find that the first weight either takes the value of 1, i.e. captures all the weight, or it takes a smaller decay dependent value; see Fig. \ref{w2dec}.  Fig. \ref{fracs}  suggests that there are two more points with vanishing step size: \one  an unstable point,  \two  a  stable point ($\approx 0.625$) (which is not taken).
\par
\begin{figure}
\psfrag{e=0.1}{$\epsilon=0.1$}
\psfrag{e=0.0005}{$\epsilon=0.0005$}
\psfrag{e=0.00001}{$\epsilon=0.00001$}
\centering
\includegraphics[angle=-90,width=0.75\textwidth]{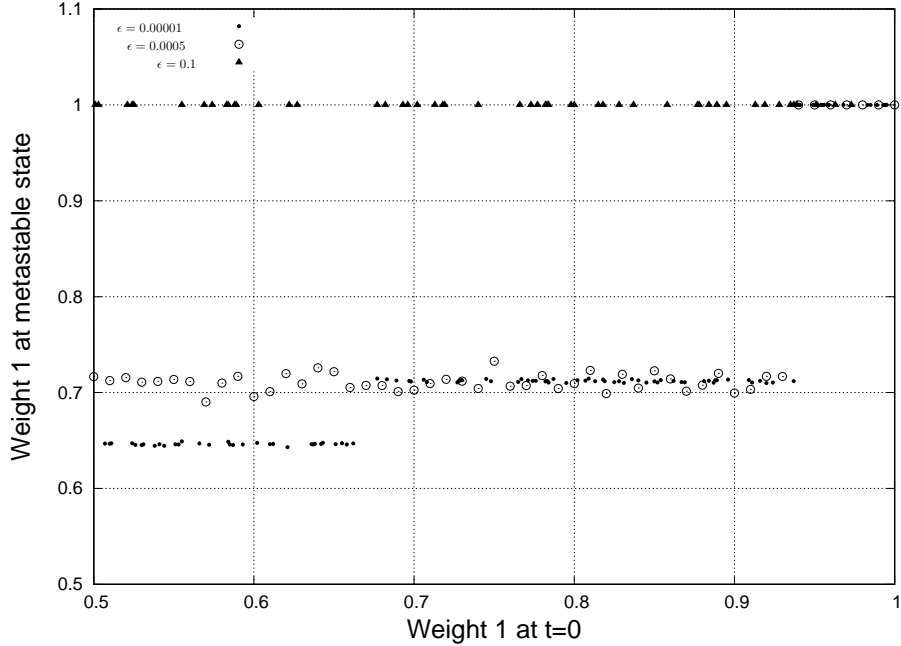} \label{wd01}
 \caption{As Fig. \protect\ref{w2dec}, but now  showing the case of  $d=0.1$ only  for 3 different step sizes. The intermediate step size is  the same data as in Fig. \protect\ref{w2dec}. The lowest step size $\epsilon=0.00001$ occupies and additional metastable state. The largest step size (implemented as a uniformly distributed, random number smaller than $0.1$) reaches the absorbing state from all initial weights.}
 \label{example2}
 \end{figure} 
\par
In order to understand this, some subtleties   need to be taken into account in the interpretation of the results.  The evolution of weights over time  is fundamentally  a random walk in weight space, rather than a deterministic dynamical system. In a stochastic system, the stable points are not actually stable, in the sense of dynamical systems theory, but rather they are points that can capture the dynamics {\em for some time}. These points  are thus better described as {\em metastable states}.  In principle escape from these metastable states is always possible. The escape rate depends on the step size (i.e. the learning rate) and the size of the basin of attraction. Note that in addition to the metastable states, the current setup also has absorbing-states from which no escape is  possible. These are  $w_1=1$ and $w_1=0$, respectively.
\par
 We can therefore expect  the system to occupy  different meta-stable states  for different amounts of time  depending on the size of the basin of attraction  of the metastable point and the   step size chosen. Larger step sizes lead to smaller dwell times of the system in the neighbourhood of metastable states. 
\par
This is illustrated   in  Fig. \ref{example2} which shows the metastable states as a function of the starting values for 3 different step sizes and a decay rate of $d=0.1$.  Of the three considered step sizes, only the smallest one occupies the  metastable state at $\approx 0.625$; in contrast,  when we chose a very large step-size, then the system always converged to the absorbing state. In-between these two extreme cases, the Hebbian learning drove the weights to two different metastable states, depending on the initial condition.  The learning rate is thus not merely a parameter that determines how fast metastable states are approached, but also which ones.

\subsubsection{Biased inputs}
In the previous examples, the input  spikes  of the two channels are assumed to be  independent, identically distributed. Our derivations did not use this assumption, hence the  results should not depend on it. To check this, we biased inputs to channel  1 relative to channel  2, while keeping the combined frequency of firing constant. For example, a bias of $0.6$ means that 60\% of input signals come via channel 1. 
\par
\begin{figure}
\centering
\subfloat[][]{\includegraphics[angle=-90,width=0.45\textwidth]{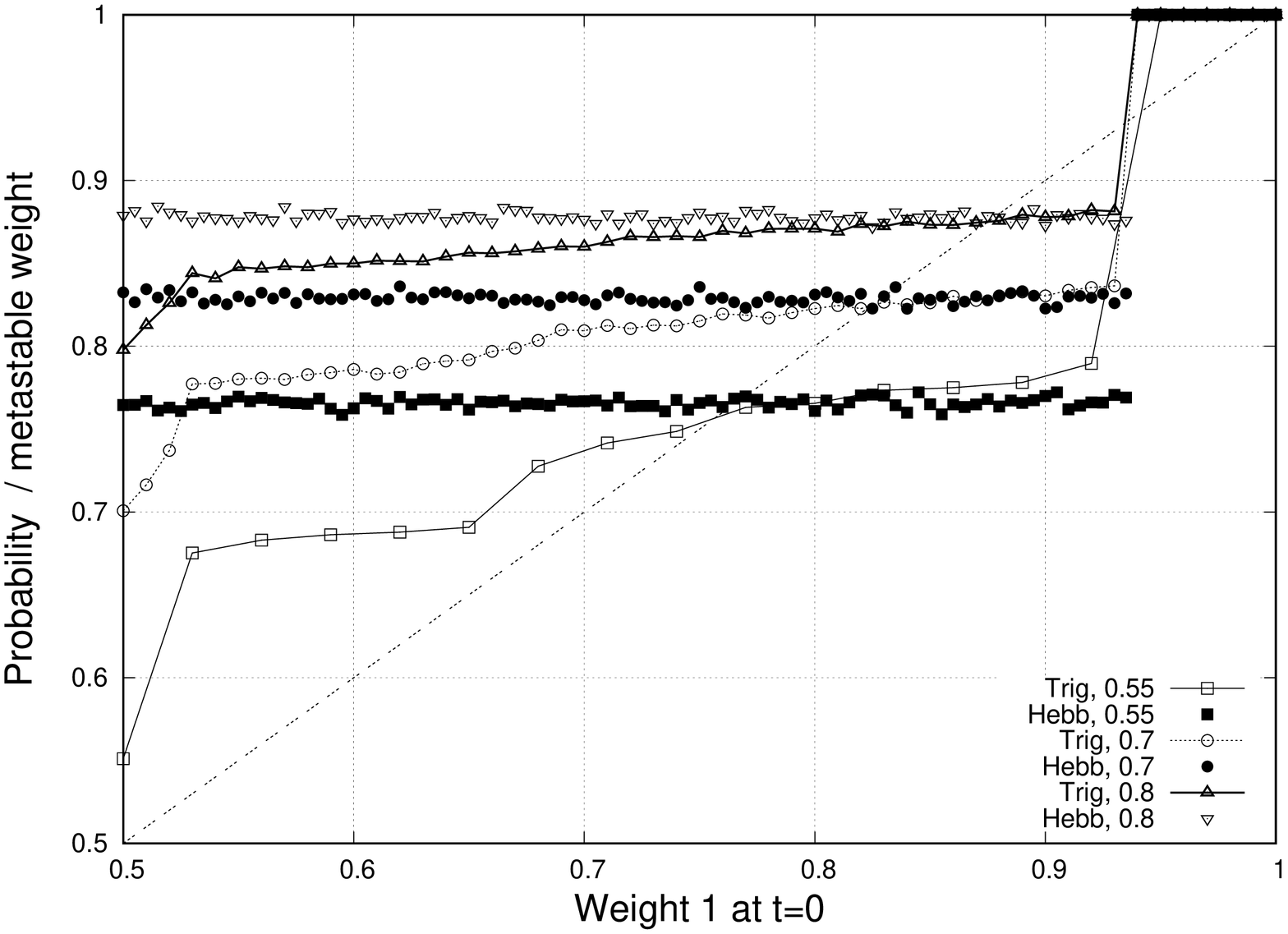} \label{examplebias}}
\subfloat[][]{\includegraphics[angle=-90,width=0.45\textwidth]{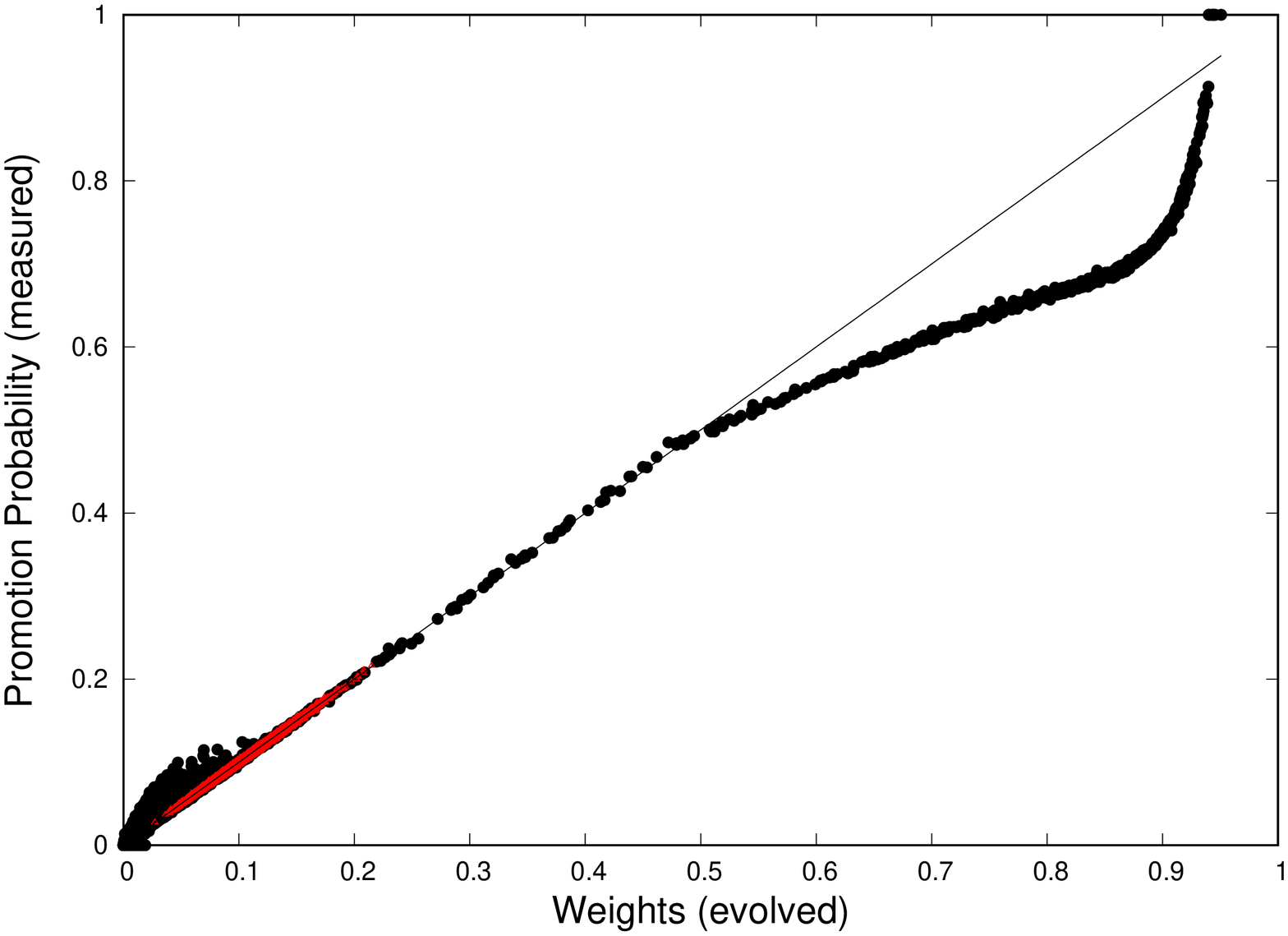} \label{examplebias10}}
 \caption{\protect\subref{examplebias} Example simulations for the case of $d=0.1$ for three different biases again from a neuron with 2 input channels. Here, the probability that input 1 fires is $0.55, 0.7$ and $0.8$ respectively. The points labelled ``Trig'' show the probability that the corresponding weight triggers an output spike calculated exactly. The points labelled  ``Hebb'' show the final weight reached by Hebbian training as a function of the initial weight.
\protect\subref{examplebias10} The black dots show the relationship between $p_i$ and the weight for a model with 10 input channels no modification of weights. The red points represent weights evolved  using Hebbian learning with $l=1,\uptau=0.15$. The frequency of input channels is  fixed but randomly chosen  from a  Gaussian distribution with a standard deviation of $0.15$ and a mean of $0.9$, i.e. all ten input channels had slightly different frequencies.}
 \label{examplebiasgraph}
\end{figure}

Fig. \ref{examplebias} shows  $p_1$  vs $w_1$ for various biases. It compares simulations where Hebbian learning is switched on with a simulation where weights are initialised randomly and then kept constant. The piecewise constant shape of the evolved weights vs the starting weights confirms that the trained values take only 1 of two  possible values (up to some noise). Furthermore, if the prediction of Eq. \ref{simplestcase} is correct, then the trained weights should take values where the promotion probability equals the normalised weights. This means that the trained and untrained curves should intersect at the diagonal, which they do in Fig. \ref{examplebias}. Thus, as expected the probabilities $p_i$ depend on the input data, but the constraint of Eq. \ref{gennorm} has to be fulfilled regardless. 

\subsubsection{More than 2 weights}

Next, we consider an example of a neuron with 10 inputs and  non-iid inputs. In this case, it is no longer practical to calculate the promotion probabilities  exactly. Instead, we need to determine it from a full simulation of the neuron.  The example simulation in Fig. \ref{examplebias10} shows that the value of the evolved weights remains bound  to those values where the promotion probability is equal to the weights.

\subsubsection{Novelty detection}

\begin{figure}
 \subfloat[][]{ \includegraphics[angle=-90,width=0.9\textwidth]{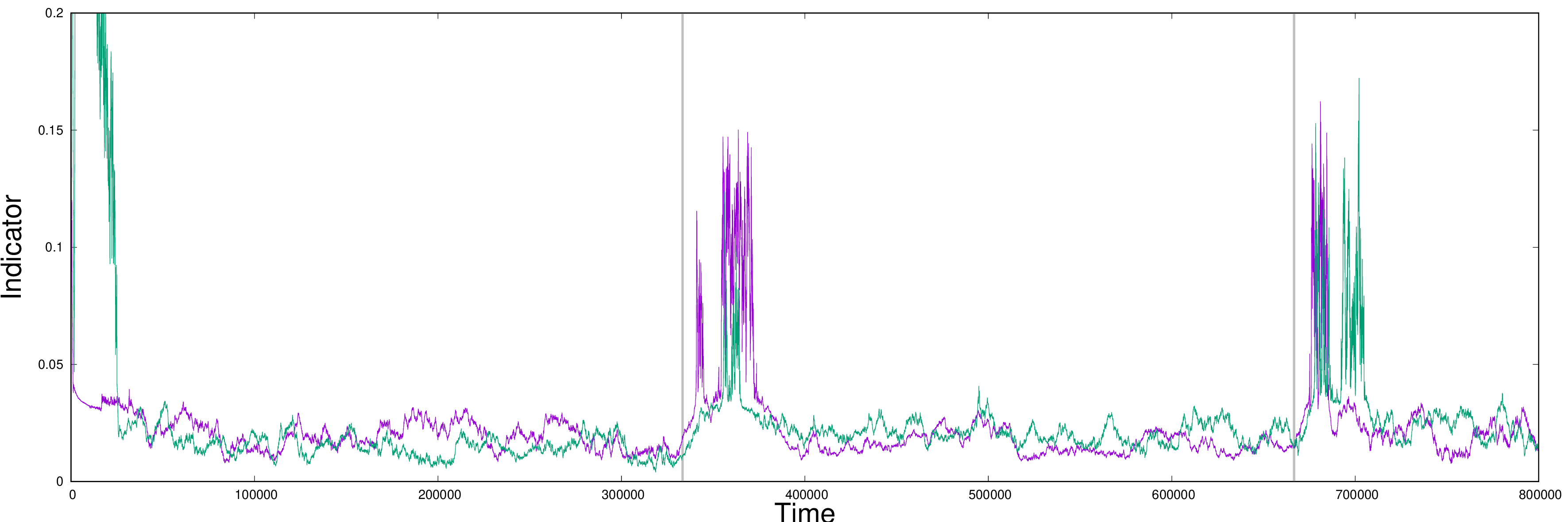}  \label{novel} }\\
 \subfloat[][]{ \includegraphics[angle=-90,width=0.9\textwidth]{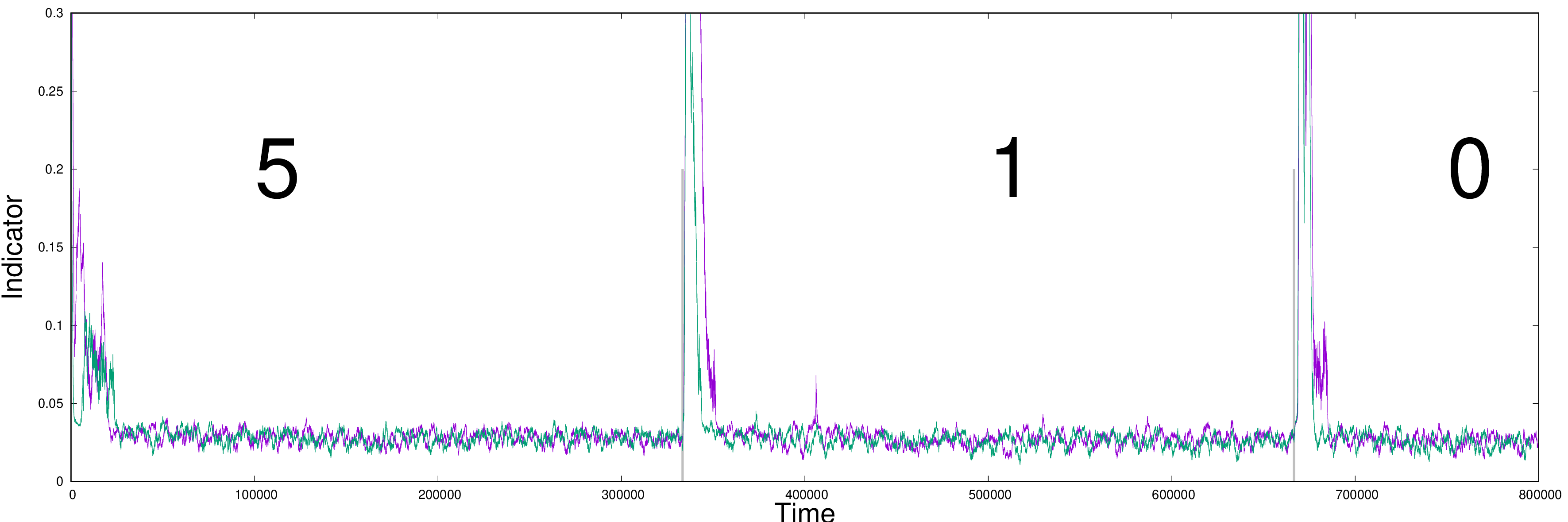} \label{mnist}}
 \caption{Distance from steady state as measured by $\varDelta$.\protect\subref{novel}  The neuron has 10 input channels and  is initialised with random weights. Each input channel fires randomly with a frequency drawn from a Gaussian distribution with mean $\mu=0.9,\sigma=1$. At times 333333 and 666666 (indicated by the grey bars) the input  frequencies are re-drawn. The graph shows two different simulations. At the time of the initialisation  Eq. \protect\ref{epsilonconstant} does not hold, but after some time the weights reach steady state and the difference approaches 0. Clearly, the change of input statistics results in a transient spike, indicating a large deviation from Eq. \protect\ref{epsilonconstant}. To prevent random deviations of the neuron from its steady state value, we made the learning rate dependent on the value of $\varDelta$ via  $\epsilon= \min(0.001,\exp(-1/4\varDelta))$. \protect\subref{mnist}  Same, but now data from the MNIST data-set. We  show the neuron images of the digits, 1, 5 and 0. See main text for detailed explanation.}
\label{neu}
 \end{figure} 
Eq. \ref{epsilonconstant} makes it possible to estimate how far the weights of the neuron are from their steady state.   We reproduce this equation here:
\begin{equation}
\underbrace{
{k^\mathrm{p}p_i(1 - w^{l-1}_i\epsilon) 
-
k^\mathrm{d}q_i(1 + w^{l-1}_i\epsilon) 
\over
k^\mathrm{p}(1-\left\langle w^{l-1} \right\rangle_\mathrm{p} )
-
k^\mathrm{d}(1+\left\langle  w^{l-1}   \right\rangle_\mathrm{d} )
}
}_{=:L_i}
   =
{ w_i\over \sum_j w_j}    
\tag{\protect\ref{epsilonconstant}}
\end{equation}
From this we can now define the  indicator 
\begin{displaymath}
\varDelta :=\sum_i\left|  L_i-{ w_i\over \sum_j w_j}\right|.
\end{displaymath}
Note that this indicator cannot be negative.  In the ideal case of the neuron being perfectly adapted, we would expect from Eq. \ref{epsilonconstant} that the indicator $\varDelta =0$. For finite learning rates, however, the weights will tend to be perturbed away from the steady state, such that in reality $\varDelta \geq 0$.  It is possible to monitor $\varDelta$ continuously during  the operation of the neuron, which in turn makes it possible to estimate how far the weights are from a steady state.   In steady state the value of $\varDelta$ will fluctuate around some small, but positive value.  Away from steady state, Eq. \ref{epsilonconstant} no longer holds; $\varDelta$ will then deviate substantially from 0.  It is thus possible to use  the value of $\varDelta$  to check for changes in the input statistics of a neuron. This turns the spiking neuron into a novelty detector.
\par
To demonstrate this, we initialised a neuron with 10 input channels to random weights. Then we provided input with  firing rates that are randomly distributed around $0.9$, as in Fig. \ref{stdpexample}. We let the neuron adjust its weights for 333333 time units, at which point we draw new random frequencies for all the channels. At time 666666 we do the same again. Throughout the simulation, we monitor the value of $\varDelta$.  Fig. \ref{novel} shows two  trajectories of $\varDelta$, which differ only by the random seed. As expected, following initialisation $\varDelta$ descends quickly to a value close to 0, where it remains untill the input statistics changes (indicated by grey bars in Fig. \ref{novel}); subsequent to these changes of inputs   there is a  substantial, but transient, deviation of $\varDelta$ from its steady state value. We conclude, that the indicator $\varDelta$ can be used for novelty detection. 
\par
Next, we tested novelty detection on the MNIST dataset of hand-written digits \cite{mnist} to ensure that it works on real-world data as well. The images of the MNIST datasets have $28\time 28$ pixels each with a greyscale value. This format is not immediately suitable for processing by a spiking neuron. We therefore pre-processed the images from the MNIST as follows:  Firstly, rather than using the entire image (consisting of 784 pixels), we limited ourselves to  the horizontal line 14 of the image. This is merely to speed up the computation.  With this, we are then left  for each image with a  vector of intensities $I:= [I_i,I_2,\ldots,I_{28}]$ representing  the intensities of each pixel in the line. We convert this vector into a single spike as follows:
\begin{enumerate}
 \item 
Compute the normalised intensity vector, $\bar I={I\over \sum_{i=1}^{28} I_i}$. 
\item
Draw a single random index $j$ of $\bar I$ with probability $p(j)= \bar I_j$.
\item
Provide an input spike to channel $j$ of the neuron. 
\end{enumerate} 
\par
The entire procedure we followed is as follows: We initialised  a neuron with 28 input channels. We then generated firing times such that the neuron as a whole receives input with a frequency of $0.9\times 28$.   For each scheduled input spike we then  chose a randomly  drawn image of the MNIST dataset; based on this image, we then chose an input channel as outline in the previous paragraph; a spike is then provided into this channel at the given time.
\par
In the experiments we performed, we started by  presenting only images drawn from the set of  images labelled as  digit 5. After 333333 time units, we then showed images of the digit 1, followed by the digit 0 after time 666666. Each change of image was marked by a spike of the value of the  indicator $\varDelta$; see Fig. \ref{mnist}. This demonstrates that the neuron can act as a novelty detector for classes of images.

\section{Discussion} 

In this article we derived constraints  on the weights of a continuous time spiking neuron, trained using Hebbian learning.   Perhaps the most surprising conclusion we found is apparent from Eq. \ref{symmcase}.  Irrespective of  the normalisation rule one uses, the weights resulting from Hebbian learning rules are constrained with respect to the normalised weights, that is weight normalisation to 1.  The actual exponent used for the normalisation, in contrast, appears only as a higher order correction.
\par
In the case of pure Hebbian learning  we find  that (up to corrections of first order) the normalised weights equal the promotion probability. The same will not be generically true for any set of weights, such that this relation can be used to characterise weights evolved by  Hebbian learning. 
\par
 In the case of asymmetric  STDP the constraints on the weights become more complicated.  In this case, the evolved weights have to be in a relationship with both the demotion and the promotion probabilities.   This  is best seen  by Eq. \ref{epsilonconstant}, which is reproduced here in the zero-order version for clarity:
\begin{equation}
{w_i\over \sum_j w_j} 
= {k_\mathrm{p} p_i - k_\mathrm{d}q_i
\over
 k_\mathrm{p}  - k_\mathrm{d}
}
\end{equation}
This tells us that the normalised weights equal  the  difference between the {\em unconditional}  promotion probability and the {\em unconditional} demotion probability divided by the difference between the total promotion and demotion probability for all weights. Note that the normalisation exponent does not appear in the zero order approximation that we reproduced here. This means that  in lowest order the weights have to fulfil the same conditions irrespective of the exponent used for normalisation.  Given that the values of weights will be smaller or equal to 1, the correction terms  will be  bounded by  $1 \pm \epsilon$. To illustrate this, when using a  learning rate of $0.0001$ then  the correction to the second term is at most $1.0001$ and at least $0.999$, but could be much closer to 1.  We hasten to stress that our results do not imply that there is no difference in the dynamics of weight evolution between various normalisation schemes. Yet, when it comes to the constraints that are fulfilled by  a metastable set of weight the exponent does not matter. 
\par
The relationships we derived are surprisingly simple and valid across a wide range of assumptions. This makes them intrinsically interesting, while also providing some novel insights into the kinds of weights one obtains from Hebbian learning. Apart from this, they are also  useful practically. Since relations such as Eq. \ref{epsilonconstant} are only valid in steady state, they can be used to monitor the distance of the weights from steady state. An immediate application of this is novelty detection, as we demonstrated above. Another possible use of this is to make the learning rate dependent on the distance from the steady state. In this way, it is possible, for example, to freeze the behaviour of the neuron as it approaches its steady state. 
\par
Our results also provide an interesting perspective on the step size $\epsilon$, which acts as the learning rate but doubles up as a correction parameter in approximations.  The normal intuition in those situations is that, as the correction parameter is decreased, the system becomes more and more similar to a specific zero-order solution.  
\par
This is the case here as well. As far as the corrections are concerned the step size only leads to a small correction to the zero-order equation, as discussed above. There is, however, an additional subtlety.    At the same time as the   learning rate is only a  correction term to the constraints, it may have a major impact  on the solutions that can be found within reasonable times. For the case of a minimal neuron with only two input channels this effect is illustrated above in Fig. \ref{example2}. The insight remains valid for neurons of any number of input channels.
 Small learning rates will tend to get stuck in the first possible combination of solutions that meet the steady state conditions, whereas for larger learning rates the  neuron will  settle on metastable states with larger basins of attraction.
In this context, it is also interesting to recall that in search-based optimisation, the learning rate is thought to impact the speed  of convergence of the algorithm as well as the accuracy with which any optimal value can be approached. In contrast, here we found that the learning rate can impact on the type of meta-stable behaviour that the system occupies.
\par
In this paper we considered mostly the case of randomly firing input channels. This allowed us to illustrate the behaviour of the learning rules in absence of any effects that are specific to particular datasets or choices on the input statistics.  The equations we used do not contain any explicit assumptions about the nature of the input frequencies, in particular they do not rely on the fact that we chose random/unbiased input.  The constraints we derived, therefore, remain valid, irrespective of input data; see also Fig. \ref{examplebias10} for an example. However, note that  the conditional promotion and demotion probabilities are a function of the weights and a function of the input data. Therefore, if the input data changes, then the conditional probabilities change. 
\par
The results presented here are exclusively about a single neuron. It will be interesting to extend this to networks of neurons. Clearly, then the constraints have to be true for each individual component of the network. We leave it to future work to explore the consequences of this for the possible weight configurations of  a spiking neural network.

%\bibliographystyle{unsrt}
%\bibliography{../../bibl}

\end{document}